\documentclass{article}
\usepackage{spconf,amsmath,graphicx}
\usepackage{algorithm}
\usepackage{amsmath,mathtools}
\usepackage{amssymb}
\usepackage{xcolor}
\usepackage{algorithm, algorithmic}
\usepackage{multirow}
\usepackage{textcomp,booktabs}
\usepackage{color}
\usepackage{colortbl}
\usepackage[hyphens]{url}  
\usepackage{subfigure}
\definecolor{mygray}{gray}{.9}
\usepackage{algorithmic}
\usepackage{caption} 

\title{Generic Dependency Modeling for Multi-Party Conversation}
\name{Weizhou Shen, Xiaojun Quan\thanks{Corresponding author: Xiaojun Quan,}\thanks {Email: quanxj3@mail.sysu.edu.cn.}, Ke Yang}
\address{Sun Yat-sen University}
\begin{document}
%
\maketitle
\begin{abstract}
To model the dependencies between utterances in multi-party conversations, we propose a simple and generic framework based on the dependency parsing results of utterances. Particularly, we present an approach to encoding the dependencies in the form of relative dependency encoding (ReDE) and illustrate how to implement it in Transformers by modifying the computation of self-attention. Experimental results on four multi-party conversation benchmarks show that this framework successfully boosts the general performance of two Transformer-based language models and leads to comparable or even superior performance compared to the state-of-the-art methods. The codes are available at \url{https://github.com/shenwzh3/ReDE}. 
\end{abstract}
\section{Introduction}

\begin{figure}[t]
  \centering
  \includegraphics[width=0.98\linewidth]{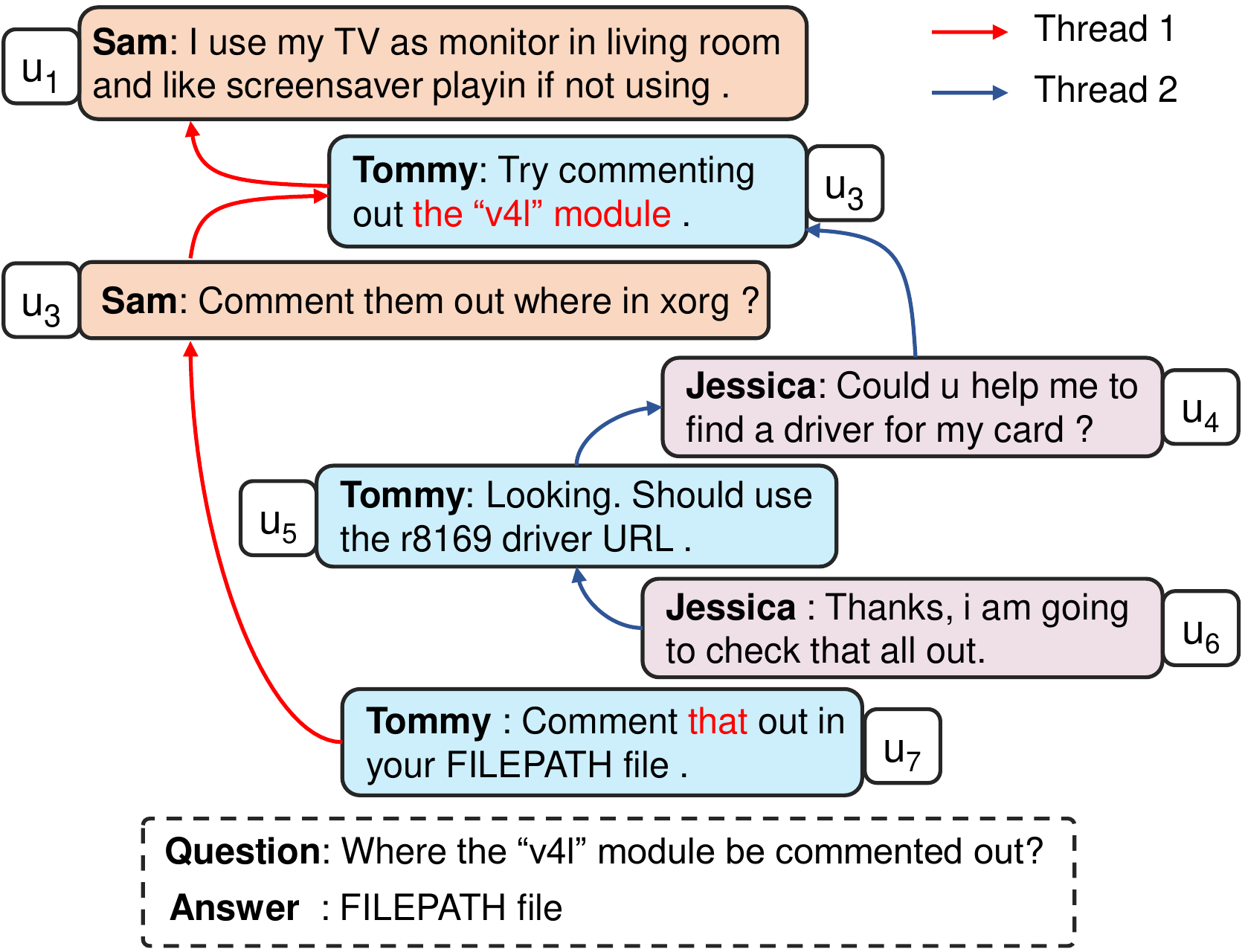}
  \caption{An example to show that utterance dependencies should be modeled in multi-party conversations. }
  \label{fig:introduction}
  \vspace{-0.5cm}
\end{figure}

Most current research on dialog systems focuses on interactions between two interlocutors. There has been a strong need for extending it to multi-party conversations since numerous scenarios naturally involve more than two interlocutors. Yet it is more challenging to model multi-party conversations as there can be complicated dependencies existing between utterances. As illustrated in Figure \ref{fig:introduction}, multi-party conversations often face the challenge of referential ambiguity and may lack coherence between consecutive utterances~\cite{li2020molweni}. This implies that it could be beneficial to take into account structural information when processing multi-party conversations.~One very intuitive idea is to encode utterance dependencies generated from heuristic rules~\cite{ijcai2019-0696} or utterance dependency parsing~\cite{shi2019deep}.  

It is natural to model utterance dependency parsing results with the well-recognized Transformers \cite{vaswani2017attention}. Two common implementations are hierarchical methods and utterance masking methods. Hierarchical methods  \cite{ijcai2019-0696,shen-etal-2021-directed} first encode each utterance separately and then feed the utterance representation into a graph neural network built according to the utterance dependencies. These methods are restricted to tasks that demand only utterance-level representation and may fall short in dealing with tasks that require token representation such as reading comprehension. On the other hand, utterance masking methods~\cite{nguyen2019tree,liu2021filling} directly mask self-attention weights between utterances that have no dependency relation. However, these methods are insensitive to different parsing results and may not capture the dependencies as desired.

We propose a simple framework that aims to endow Transformers with the genericity to model utterance dependencies. We first acquire the utterance dependencies in a multi-party conversation with an off-the-shelf utterance dependency parser~\cite{shi2019deep}. To encode the utterance dependencies, we propose \emph{relative dependency encoding} (ReDE) that encodes the relative dependency distance of two utterances. The encoded dependency information is incorporated into Transformers by modifying the computation of self-attention. Without adding a new module, this framework can be easily applied to a wide range of Transformer-based language models and multi-party conversation tasks. Besides, we introduce a simple way to initialize the dependency-related parameters to further enhance the impact of utterance dependencies in self-attention and fine-tune the models more easily.

We evaluate the new framework with two pre-trained language models, RoBERTa \cite{liu2019roberta} and BART \cite{lewis2020bart} on four multi-party conversation benchmarks, including emotion detection, relation extraction, machine reading comprehension, and abstractive summarization. Results show that this framework successfully boosts the performance of the two models on all the tasks, verifying its effectiveness and genericity.

\section{Methodology}

\subsection{Formulation}
A multi-party conversation can be formulated as a sequence $\mathcal{U} = \{u_0, u_1, ..., u_{N-1}\}$ of utterances, in which each utterance $u_i$ comprises a sequence of tokens uttered by a speaker $s(u_i)$. 
We concatenate the utterances and define an indicator function $I(i)$ to map the $i$th token to its utterance. 

As illustrated in Figure \ref{fig:introduction}, utterance dependencies provide structural information of some kind in multi-party conversations and can be characterized by linguistic relations from one utterance to another. 
For the sake of simplicity, we only consider the connectivity of two utterances in this work and leave their relation types to future research. Formally, the utterance dependencies for a conversation are represented by a set $\mathcal{E}$ of directed relations in the form of $u_i \rightarrow u_j$. To extract these relations, an utterance dependency parser is applied:
\begin{equation}
    \mathcal{E} = \text{Parse}(\{u_0, u_1, ..., u_{N-1}\}).
\end{equation}

The dependency parser forces each utterance to point to exactly one of its preceding utterances in the same conversation. Therefore, the parsing result of the conversation together with the utterances naturally forms a tree rooted at node $u_0$.

We encode the utterance dependencies in multi-party conversations at the token level in each Transformer self-attention computation. Specifically, for each attention head in the Transformer layers, the attention score between the $i$th and $j$th tokens in the input sequence is computed as:
\begin{equation}\label{eq:self-attn}
    s_{ij} = \frac{x_i W_q W_k^{\top} x^{\top}_j + a_{ij}}{\sqrt{d}},
\end{equation}
where $x_i W_q W_k^{\top} x^{\top}_j$ is the token-level self-attention term in Transformer~\cite{vaswani2017attention}, $x_i$ and $x_j$ are the token representations, and $a_{ij}$ is our utterance dependency bias term to represent the degree of utterance dependency between the $i$th and $j$th tokens. 



\begin{figure}[t]
  \centering
  \includegraphics[width=0.8\linewidth]{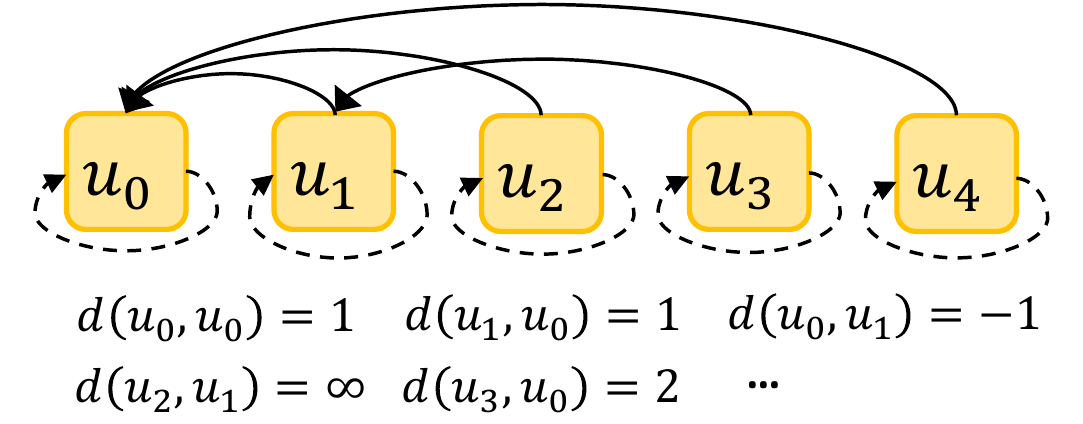}
  \caption{Demonstration of utterance dependency encoding}
  \label{fig:dep_enc_sub_rel}
  \vspace{-0.3cm}
\end{figure}

\subsection{ReDE: Relative Dependency Encoding}\label{sec:rel_enc}

Intuitively, utterances with a shorter dependency distance should be semantically close to each other. Therefore, we take the distance between two utterances in the dependency tree as input to implement relative utterance dependency encoding. 
Specifically, the distance from $u_i$ to $u_j$, denoted as $d(u_i,u_j)$, is determined by the length of the path from $u_i$ to $u_j$ in the tree. We also define the reversed distance from $u_j$ to $u_i$ as the negative length of this path. If such a path does not exist, both $d(u_i,u_j)$ and $d(u_j,u_i)$ are set to $\infty$. Particularly, we define $d(u_i,u_i)=1$ for any $i$, assuming that each utterance is self-dependent. Figure \ref{fig:dep_enc_sub_rel} presents an example to illustrate how to compute the relative utterance distance. 
Moreover, if two utterances are located far away from each other in the tree, their interaction is assumably weak and can be considered as no dependency relation for the sake of simplicity. 
To this end, we clip the utterance distance to get the final relative dependency encoding:
\begin{equation}
    \hat{d}(u_i,u_j) = \left\{
\begin{array}{cl}
d(u_i,u_j),&\arrowvert d(u_i,u_j)\arrowvert \leq \tau\\
\infty,&\text{otherwise}\\
\end{array} \right.
\end{equation}
where $\tau$ is a threshold to clip the dependency distance\footnote{We set $\tau=7$ for all tasks, this value is picked from $[1,3,5,7,9]$ by hyperparameter tuning on the dev set of Molweni-MRC.}. 

The relative dependency representation between tokens $u_i$ and $u_j$ is obtained by feeding the relative dependency encoding of their utterances through a trainable embedding layer:
\begin{equation}\label{eq:emb_rel}
    r_{ij} = \text{Embed}_{r}(\hat{d}(u_{I(i)},u_{I(j)})).
\end{equation}
The attention score in Equation (\ref{eq:self-attn}) with relative utterance dependency term is re-computed as:
\begin{equation}\label{eq:attn_rel}
    s^r_{ij} = \frac{x_i W_q W_k^{\top} x^{\top}_j +  W^{r\top} r_{ij} }{\sqrt{d}},
\end{equation}
where $W^{r}$ is a trainable weight matrix.

\begin{table*}[t]
	\centering
	\resizebox{0.9\textwidth}{!}{
	\begin{tabular}{p{2.2cm}|p{1.1cm}<{\centering}|p{1.1cm}<{\centering}p{1.1cm}<{\centering}p{1.1cm}<{\centering}|p{1.1cm}<{\centering}p{1.1cm}<{\centering}|p{1.3cm}<{\centering}p{1.3cm}<{\centering}p{1.3cm}<{\centering}}
		\toprule
		\multirow{2}*{Method} & MELD&\multicolumn{3}{c|}{DialogRE} &\multicolumn{2}{c|}{Molweni-MRC}&\multicolumn{3}{c}{SAMSum}\\ 
		\cline{2-10}
		&F1 &P  &R  &F1  &EM &F1 & Rouge-1  &Rouge-2  &Rouge-L\\
		\hline
		Baseline& 63.38  & 66.88 & 61.37 & 64.00  &58.06  &71.75&  53.06   &28.35  &48.82 \\ 
		Baseline$_{hr}$& 64.61  & Failed  & Failed  & Failed  &- & -&- & -&- \\ 
		Baseline$_{mask}$& 64.21   &64.52  &53.43  &58.32 & 56.63 &71.93&53.20 	&28.31   & 48.90  \\ 
		SOTA& \textbf{66.32} & -& -&\textbf{69.10} & 58.60& 72.20& 53.43 & 28.74 & 49.19 \\
		\hline
		ReDE (ours)& 65.59    &70.39 & 66.66 &68.47  &58.20 &72.40&53.67  &28.79  &49.61 \\ 
		ReDE$_{pt}$ (ours)&66.09  & \textbf{70.49} & \textbf{67.45}  & 68.93  &\textbf{58.92}  &\textbf{73.04}&\textbf{53.94}  &\textbf{28.97} & \textbf{49.81}  \\ 
		\bottomrule
	\end{tabular}
	}
	\caption{Overall results on the four evaluated datasets.}
	\label{tab:res_discriminant}
	\vspace{-0.3cm}
\end{table*}

\subsection{Initialization of Dependency Embedding}\label{sec:initialization}
It is general to initialize the embedding vectors of Equation (\ref{eq:emb_rel}) by a Gaussian distribution $\mathcal{N}(0, 0.1)$. However, we discover that after fine-tuning the values of the bias term for utterance dependency encoding would be far smaller than the original self-attention scores, leading to a situation where utterance dependencies contribute extremely small to the computation of self-attention. The reason could be that the limited number of optimization steps during fine-tuning cannot update utterance dependency embeddings to a similar magnitude as word embeddings. Therefore, we employ a special initialization method for utterance dependency embedding to narrow the gap. Specifically, before post-training and fine-tuning, we initialize the utterance dependency embeddings by sampling vectors from a Multivariate Gaussian:
\begin{equation}
    e \sim \mathcal{N}(\mu, C),
\end{equation}
where $\mu$ is the average of all word embeddings from the original language modeling tasks, and $C$ is the covariance matrix.

We also adopt in-domain post-training of our models with a multi-party conversational corpus. To avoid forgetting the original pre-trained parameters catastrophically, we opt to update only the parameters related to utterance dependency encoding, i.e., $W_q^{a}, W_k^{a}$ and $W^{r}$, and the embedding layer of utterance dependencies, while conducting the post-training.

\section{Experimental Setup}

\subsection{Datasets}\label{sec:benchmarks}
We evaluate the proposed framework on the following four multi-party conversation benchmarks: \textbf{MELD}~\cite{poria2019meld}, \textbf{DialogRE}~\cite{yu2020dialogue}, \textbf{Molweni-MRC}~\cite{li2020molweni}, and \textbf{SAMSum}~\cite{gliwa2019samsum}, which cover the tasks of utterance-level emotion classification, relation extraction, reading comprehension, and summarization.

\subsection{Implementation Settings}
We employ the state-of-the-art utterance dependency parser Deep Sequential~\cite{shi2019deep} to parse multi-party conversations. For the discriminative tasks (MELD, DialogRE, and Molweni-MRC), we employ RoBERTa-large~\cite{liu2019roberta} as the base model. For the generative task (SAMSum), we employ BART-large~\cite{lewis2020bart}.
When implementing the two models, we first initialize them with weights released by
Huggingface\footnote{RoBERTa:~\url{https://huggingface.co/roberta-large}, BART:~\url{https://huggingface.co/facebook/bart-large}} and initialize parameters related to utterance dependencies as introduced in Section \ref{sec:initialization}. For BART-large, the utterance dependency encoding is only added to the encoder layers. 

The corpus for in-domain post-training is the combination of the four conversation datasets introduced in Section \ref{sec:benchmarks}, the STAC \cite{asher2016discourse} and Molweni-DP \cite{li2020molweni} corpora, and an open-source corpus\footnote{\url{https://github.com/emorynlp/character-mining}} collected from the TV show \emph{Friends}. For RoBERTa, the masked language modeling task is applied for in-domain post-training; and for BART, the masked language recovering task is applied.



\subsection{Compared Methods}
We employ several strong baselines for comparison, including some combined with utterance dependencies:

(1) \textbf{Baseline}~\cite{liu2019roberta,lewis2020bart} is the base model (RoBERTa / BART) directly fine-tuned on the benchmarks, without utterance dependencies. (2) \textbf{Baseline$_{hr}$} encodes each utterance with baseline and feeds the utterance representation into a directed acyclic graph network~\cite{shen-etal-2021-directed}. This hierarchical structure does not apply to the machine reading comprehension and summarization benchmarks. (3) \textbf{Baseline$_{mask}$} \cite{nguyen2019tree,liu2021filling} masks the self-attention weight of two tokens in the base model who have no dependency relation. (4) \textbf{ReDE} is the basic model with our relative utterance dependency encoding.
(5) \textbf{ReDE$_{pt}$} is \textbf{ReDE} with further in-domain post-training introduced in Section \ref{sec:initialization}.
(6) \textbf{SOTA}~\cite{chudasama2022m2fnet,qiu2021socaog,ma2021enhanced,lewis2020bart} are the state-of-the-art methods on the four evaluated datasets.

\vspace{-0.3cm}

\section{Results and Analysis}
\vspace{-0.3cm}

\subsection{Overall results}
The overall results on the four datasets are reported in Table \ref{tab:res_discriminant}, from which we can draw three observations. 
First, with a simple and generic framework, our models can reach comparable performance with the state-of-the-art methods on MELD and DialogRE, and surpass the SOTA on Molweni-MRC and SAMSum.
Second, after incorporating relative utterance dependency encoding (ReDE), the performance of Baseline improves consistently, outperforming the hierarchical method (Baseline$_{hr}$) and the masking method (Baseline$_{mask}$). This confirms the effectiveness of our proposed method. 
Finally, after in-domain post-training, further performance improvement can be seen on both ReDE$_{pt}$, but the performance gain is not as significant as that of modifying from Baseline to ReDE.

\begin{figure*}[t]
	\centering
	\includegraphics[width=0.9\textwidth]{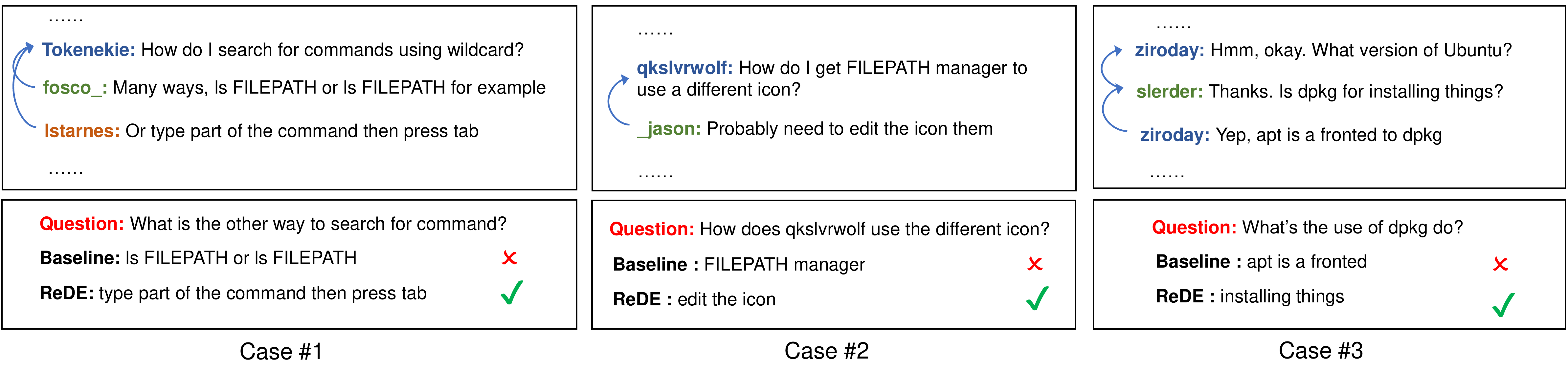}
	\caption{Three test cases in Molweni-MRC.}
	\label{fig:casestudy}
	\vspace{-0.3cm}
\end{figure*} 

\subsection{Ablation Study}\label{sec:exp_ablation}
We compare initializing the dependency-related parameters as introduced in Section \ref{sec:initialization} with the traditional method of initializing randomly from $\mathcal{N}(0, 0.1)$ (ReDE  w/o $init$). As shown in the first two lines of Table \ref{tab:initialization}, with our initialization method, ReDE achieve considerable performance gains. This means that by initializing dependency-related parameters to a similar magnitude as word embeddings, the models can be fine-tuned easier, and the rule of utterance dependency can be better highlighted, which leads to a performance gain.

We also compare the strategy of updating all the model parameters (ReDE$_{pt}$ w/o $partial$) and that of updating only the dependency-related parameters during in-domain post-training. We observe from the results in the last two lines of Table \ref{tab:initialization} that when updating all the model parameters, the performance is even worse than that without post-training. 

Since ReDE assigns the same dependency encoding for token pairs within the same utterance pair, it also plays a certain role in segmenting the input conversation into utterances. Therefore, we investigate whether the performance improvement of our method truly stems from the utterance dependencies or just because it segments the conversation. We compare ReDE with the segment encoding method introduced by Chen et al.~\cite{chen2021demystifying}. To make a fair comparison, the numbers of parameters in the compared methods are kept the same. 

We note from Table \ref{tab:res_seg} that the segment encoding method (Baseline$_{seg}$) can also boost the performance of Baseline, while our ReDE outperforms Baseline$_{seg}$. This indicates that Transformer-based models equipped with our utterance dependency encoding can obtain a more comprehensive insight into conversation than merely understanding the segments.

\begin{table}[t]
	\centering
	\resizebox{0.44\textwidth}{!}{
	\begin{tabular}{l|ccc}
		\toprule
		Model\ \ \  & MELD & DialogRE & Molweni-MRC \\ 
		\hline
		ReDE &65.59 & 68.47 &72.40\\
		\ \ \ \   w/o $init$ &64.81 & 67.15 &72.11\\
		\hline
		ReDE$_{pt}$ & 66.09 & 68.93 & 73.04\\
		\ \ \ \   w/o $partial$ & 65.44 & 68.27 & 71.85\\
		\bottomrule
	\end{tabular}
	}
	\caption{Results of F1 score on MELD, DialogRE and Molweni-MRC by different initialization methods.  }
	\label{tab:initialization}
	\vspace{-0.3cm}
\end{table}

\subsection{Case Study}

In Figure \ref{fig:casestudy}, we present three cases to show more clearly why ReDE can surpass the base model. We find that the base model tends to make reasoning only based on language patterns in conversations. For example, the question in Case \#1 is ``What is the other way ...'', baseline finds a wrong answer ``ls FILEPATH or ls FILEPATH'' in the second utterance due to the phrase ``Many ways'' can strongly match the ``other way'' in the question, and the ``...or...'' pattern to some extent has the meaning of ``the other way''. In Case \#2 the question is ``How does ... use the different icon'', baseline finds the wrong answer with the close pattern ``get ... to use a different icon'', however it fails to consider the true meaning of the selected answer. The same situation occurs in Case \#3 where the baseline obtains a wrong answer ``apt is a fronted'', because it follows the pattern ``... to dpkg'' in the context, which matches the question with the pattern ``What is the use of dpkg''. 

\begin{table}[t]
	\centering
	\resizebox{0.45\textwidth}{!}{
	\begin{tabular}{l|ccc}
		\toprule
		Methods & MELD & DialogRE & Molweni-MRC \\ 
		\hline
		Baseline &63.38 &64.00 &71.75\\
		Baseline$_{seg}$ & 64.32& 66.55& 72.19\\
		ReDE & \textbf{65.59}& \textbf{68.47}& \textbf{72.40}\\
		\bottomrule
	\end{tabular}
	}
	\caption{F1 scores on MELD, DialogRE and Molweni-MRC. RoBERTa$_{seg}$ denotes the segment encoding method \cite{chen2021demystifying}.}
	\label{tab:res_seg}
	\vspace{-0.3cm}
\end{table}

On the contrary, in the given three cases, we can generally conclude the characteristic of our proposed methods. To find an answer, ReDE can first locate the conversation snippets that are semantically related to the question, then conduct a further reasoning base on utterance dependency to find the located utterance of the potential answer, and finally yield the correct result. For example, in Case \#2, ReDE locates the utterance describing the question, and successfully find the answer in the next utterance which is in the dependency relation ship of question-answer pair with the utterance asking the question.  This indicates that our proposed methods can rectify the erroneous reasoning behavior of baseline by considering utterance dependency. These cases also show that utterance dependency is essential to solving certain tasks in multi-party conversations.

\section{Conclusion}
We present a simple and generic framework for modeling utterance dependencies to better facilitate the understanding of multi-party conversations. Particularly, we propose relative dependency encoding (ReDE) and incorporate it into Transformers by modifying the computation of self-attention. Experiments on four multi-party conversation benchmarks show this framework boosts the general performance of Transformer-based models. Through in-depth analysis, we further show that utterance dependencies are beneficial for multi-party conversation modeling and that the proposed framework can learn to model utterance dependencies effectively in various multi-party conversation tasks.

\bibliographystyle{IEEEbib}
\bibliography{strings,refs}

\begin{thebibliography}{10}

\bibitem{li2020molweni}
Jiaqi Li, Ming Liu, Min-Yen Kan, Zihao Zheng, Zekun Wang, Wenqiang Lei, Ting
  Liu, and Bing Qin,
\newblock ``Molweni: A challenge multiparty dialogues-based machine reading
  comprehension dataset with discourse structure,''
\newblock in {\em Proceedings of the 28th International Conference on
  Computational Linguistics}, 2020, pp. 2642--2652.

\bibitem{ijcai2019-0696}
Wenpeng Hu, Zhangming Chan, Bing Liu, Dongyan Zhao, Jinwen Ma, and Rui Yan,
\newblock ``Gsn: A graph-structured network for multi-party dialogues,''
\newblock in {\em Proceedings of the Twenty-Eighth International Joint
  Conference on Artificial Intelligence, {IJCAI-19}}, 7 2019, pp. 5010--5016.

\bibitem{shi2019deep}
Zhouxing Shi and Minlie Huang,
\newblock ``A deep sequential model for discourse parsing on multi-party
  dialogues,''
\newblock in {\em Proceedings of the AAAI Conference on Artificial
  Intelligence}, 2019, vol.~33, pp. 7007--7014.

\bibitem{vaswani2017attention}
Ashish Vaswani, Noam Shazeer, Niki Parmar, Jakob Uszkoreit, Llion Jones,
  Aidan~N Gomez, {\L}ukasz Kaiser, and Illia Polosukhin,
\newblock ``Attention is all you need,''
\newblock in {\em Advances in neural information processing systems}, 2017, pp.
  5998--6008.

\bibitem{shen-etal-2021-directed}
Weizhou Shen, Siyue Wu, Yunyi Yang, and Xiaojun Quan,
\newblock ``Directed acyclic graph network for conversational emotion
  recognition,''
\newblock in {\em Proceedings of the 59th Annual Meeting of the Association for
  Computational Linguistics and the 11th International Joint Conference on
  Natural Language Processing (Volume 1: Long Papers)}, Online, Aug. 2021, pp.
  1551--1560.

\bibitem{nguyen2019tree}
Xuan-Phi Nguyen, Shafiq Joty, Steven Hoi, and Richard Socher,
\newblock ``Tree-structured attention with hierarchical accumulation,''
\newblock in {\em International Conference on Learning Representations}, 2019.

\bibitem{liu2021filling}
Longxiang Liu, Zhuosheng Zhang, Hai Zhao, Xi~Zhou, and Xiang Zhou,
\newblock ``Filling the gap of utterance-aware and speaker-aware representation
  for multi-turn dialogue,''
\newblock in {\em The Thirty-Fifth AAAI Conference on Artificial Intelligence
  (AAAI-21)}, 2021, pp. 13406--13414.

\bibitem{liu2019roberta}
Yinhan {Liu}, Myle {Ott}, Naman {Goyal}, Jingfei {Du}, Mandar {Joshi}, Danqi
  {Chen}, Omer {Levy}, Mike {Lewis}, Luke {Zettlemoyer}, and Veselin
  {Stoyanov},
\newblock ``Roberta: A robustly optimized bert pretraining approach,''
\newblock {\em arXiv preprint arXiv:1907.11692}, 2019.

\bibitem{lewis2020bart}
Mike Lewis, Yinhan Liu, Naman Goyal, Marjan Ghazvininejad, Abdelrahman Mohamed,
  Omer Levy, Veselin Stoyanov, and Luke Zettlemoyer,
\newblock ``Bart: Denoising sequence-to-sequence pre-training for natural
  language generation, translation, and comprehension,''
\newblock in {\em Proceedings of the 58th Annual Meeting of the Association for
  Computational Linguistics}, 2020, pp. 7871--7880.

\bibitem{poria2019meld}
Soujanya Poria, Devamanyu Hazarika, Navonil Majumder, Gautam Naik, Erik
  Cambria, and Rada Mihalcea,
\newblock ``Meld: A multimodal multi-party dataset for emotion recognition in
  conversations,''
\newblock in {\em Proceedings of the 57th Annual Meeting of the Association for
  Computational Linguistics}, 2019, pp. 527--536.

\bibitem{yu2020dialogue}
Dian Yu, Kai Sun, Claire Cardie, and Dong Yu,
\newblock ``Dialogue-based relation extraction,''
\newblock in {\em Proceedings of the 58th Annual Meeting of the Association for
  Computational Linguistics}, 2020, pp. 4927--4940.

\bibitem{gliwa2019samsum}
Bogdan Gliwa, Iwona Mochol, Maciej Biesek, and Aleksander Wawer,
\newblock ``Samsum corpus: A human-annotated dialogue dataset for abstractive
  summarization,''
\newblock in {\em Proceedings of the 2nd Workshop on New Frontiers in
  Summarization}, 2019, pp. 70--79.

\bibitem{asher2016discourse}
Nicholas Asher, Julie Hunter, Mathieu Morey, Benamara Farah, and Stergos
  Afantenos,
\newblock ``Discourse structure and dialogue acts in multiparty dialogue: the
  stac corpus,''
\newblock in {\em Proceedings of the Tenth International Conference on Language
  Resources and Evaluation (LREC'16)}, 2016, pp. 2721--2727.

\bibitem{chudasama2022m2fnet}
Vishal Chudasama, Purbayan Kar, Ashish Gudmalwar, Nirmesh Shah, Pankaj Wasnik,
  and Naoyuki Onoe,
\newblock ``M2fnet: Multi-modal fusion network for emotion recognition in
  conversation,''
\newblock in {\em Proceedings of the IEEE/CVF Conference on Computer Vision and
  Pattern Recognition}, 2022, pp. 4652--4661.

\bibitem{qiu2021socaog}
Liang Qiu, Yuan Liang, Yizhou Zhao, Pan Lu, Baolin Peng, Zhou Yu, Ying~Nian Wu,
  and Song-chun Zhu,
\newblock ``Socaog: Incremental graph parsing for social relation inference in
  dialogues,''
\newblock in {\em Proceedings of the 59th Annual Meeting of the Association for
  Computational Linguistics and the 11th International Joint Conference on
  Natural Language Processing (Volume 1: Long Papers)}, 2021, pp. 658--670.

\bibitem{ma2021enhanced}
Xinbei Ma, Zhuosheng Zhang, and Hai Zhao,
\newblock ``Enhanced speaker-aware multi-party multi-turn dialogue
  comprehension,''
\newblock {\em arXiv preprint arXiv:2109.04066}, 2021.

\bibitem{chen2021demystifying}
Pu-Chin Chen, Henry Tsai, Srinadh Bhojanapalli, Hyung~Won Chung, Yin-Wen Chang,
  and Chun-Sung Ferng,
\newblock ``Demystifying the better performance of position encoding variants
  for transformer,''
\newblock {\em arXiv preprint arXiv:2104.08698}, 2021.

\end{thebibliography}

\end{document}